\documentclass{article}

\usepackage{microtype}
\usepackage{graphicx}
\usepackage{booktabs} 
\usepackage{threeparttable}
\usepackage{amsmath}
\usepackage{multirow}
\usepackage{svg}

\usepackage{hyperref}

\usepackage[accepted]{icml2021}
\usepackage{subcaption}
\usepackage{pifont}
\usepackage{xcolor}
\definecolor{pastelgreen}{rgb}{0.5, 0.8, 0.5}
\definecolor{pastelred}{rgb}{0.8, 0.31, 0.36}
\newcommand{\cmark}{\color{pastelgreen}\ding{51}}
\newcommand{\xmark}{\color{pastelred}\ding{55}}

\icmltitlerunning{Probabilistic Embeddings Revisited}

\begin{document}

\twocolumn[
\icmltitle{Probabilistic Embeddings Revisited}



\icmlsetsymbol{equal}{*}

\begin{icmlauthorlist}
\icmlauthor{Ivan Karpukhin}{tinkoff}
\icmlauthor{Stanislav Dereka}{tinkoff}
\icmlauthor{Sergey Kolesnikov}{tinkoff}
\end{icmlauthorlist}

\icmlaffiliation{tinkoff}{Tinkoff}

\icmlcorrespondingauthor{Ivan Karpukhin}{i.a.karpukhin@tinkoff.ru}

\icmlkeywords{Deep Learning, Representation Learning, Metric Learning, Confidence Estimation}

\vskip 0.3in
]



\printAffiliationsAndNotice{}  

\begin{abstract}
In recent years, deep metric learning and its probabilistic extensions claimed state-of-the-art results in the face verification task. Despite improvements in face verification, probabilistic methods received little attention in the research community and practical applications.

In this paper, we, for the first time, perform an in-depth analysis of known probabilistic methods in verification and retrieval tasks. We study different design choices and propose a simple extension, achieving new state-of-the-art results among probabilistic methods.
Finally, we study confidence prediction and show that it correlates with data quality, but contains little information about prediction error probability.
We thus provide a new confidence evaluation benchmark and establish a baseline for future confidence prediction research. PyTorch implementation is publicly released.

\end{abstract}

\section{Introduction}
\begin{figure}[t!]
\vskip 0.3in
\centering
\begin{subfigure}[b]{.4\columnwidth}
\hspace*{3mm}
\includegraphics[width=76mm]{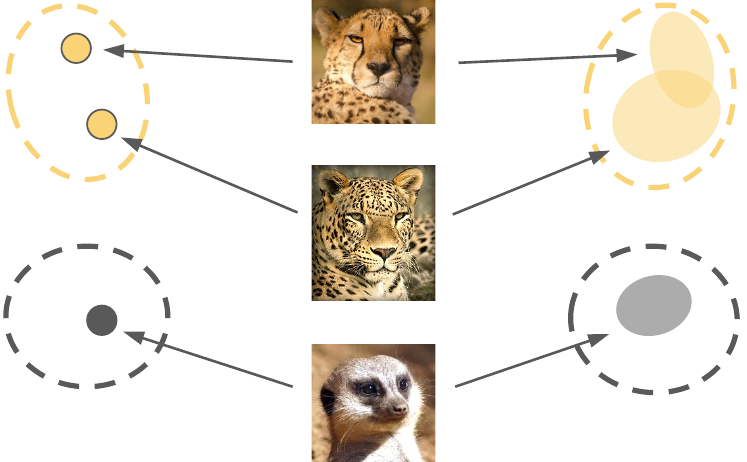}
\vskip -0.2in
\centering
$z = f(x)$
\subcaption{Deterministic}
\end{subfigure}
\hfill
\begin{subfigure}[b]{.4\columnwidth}
\centering
$\mathrm{P}(z \vert x)$
\subcaption{Probabilistic}
\end{subfigure}
\caption{(a) Deterministic methods map input image to a single embedding vector. (b) Probabilistic methods predict the distribution of embeddings. Same-class embeddings are close to each other while embeddings from different classes are far away.
}
\label{fig:pml}
\end{figure}

The goal of metric learning is to map data to an embedding space so that the embeddings of similar data are close together and those of dissimilar data are far apart \cite{bengio2013representation,musgrave2020metric}.
Deep metric learning achieved state-of-the-art results in many computer vision tasks, including face verification \cite{wang2020facerec, huang2014lfw}, person re-identification \cite{cheng2016tripletreid}, and image retrieval \cite{huang2015tripletretrieval, oh2016sop}.

Motivated by previous research in mixture density networks \cite{bishop1994mixture} and variational auto-\mbox{encoders \cite{kingma2014vae}}, probabilistic extensions were proposed to deep metric learning \cite{oh2018hib, shi2019probabilistic,scott2019spe,chang2020dul, li2021spherical, scott2021vmfloss}. Within this approach, which we call {\it probabilistic embeddings} (PE), the algorithm predicts a distribution of embeddings rather than a single vector, as shown in Figure \ref{fig:pml}.
Compared to traditional metric learning algorithms, which we call {\it deterministic methods}, PE have the following advantages: (1) probabilistic losses can stabilize training on noisy data \cite{oh2018hib}, (2) PE allows accurate embeddings aggregation and comparison \cite{shi2019probabilistic, scott2021vmfloss}, (3) predicted uncertainty can be used to measure data quality and detect out-of-domain data \cite{scott2019spe}. Confidence can potentially be applied for classification with rejection tasks \cite{mena2020uncertainty, shi2019probabilistic, chang2020dul}. However, to the best of our knowledge, there are currently no works confirming that confidence is accurate enough for robust error detection.

While PE methods claimed quality improvements in face verification \cite{chang2020dul, li2021spherical} and handwritten character recognition tasks \cite{oh2018hib, scott2019spe}, they received little attention in the research community. Recent metric learning benchmarks avoided comparison with PE \cite{musgrave2020metric, roth2020revisiting}, and is still unclear how different PE methods compare to each other and how they perform in different retrieval tasks.

In this work, we perform a comparison and in-depth analysis of probabilistic embeddings.
The main contributions of this paper can be summarized as follows:
\begin{enumerate}
    \item We propose an evaluation protocol and compare known probabilistic embeddings approaches with modern deterministic methods on the following datasets: CUB200 \cite{welinder2010cub}, Cars196 \cite{krause2013cars}, In-shop Clothes Retrieval Benchmark \cite{liu2016inshop}, and Stanford Online Products \cite{oh2016sop}. We focus on both verification and retrieval quality. According to our results, probabilistic methods improve retrieval on datasets with thousands of classes (In-shop, SOP), while there is no clear improvement on datasets with hundreds of classes (CUB200, Cars196).
    \item We perform ablation studies which were not found in previous works. In particular, we studied the effects of distribution type and comparison function on verification and retrieval quality. According to our experiments, simple techniques such as normal distribution and cosine similarity achieve on-par or slightly better results than more sophisticated approaches.
    \item In our qualitative and quantitative studies, we show that confidence, predicted by probabilistic methods, can be used for data quality estimation. In particular, predicted confidence better correlates with the degree of data corruption than simple non-probabilistic baselines. We also show that confidence, predicted by current methods, produces only a small improvement in retrieval on rejection tasks, compared with trivial approaches.
    \item We propose an improved probabilistic method, named DUL-reg-cls, which outperforms previous probabilistic approaches in most comparisons. We therefore claim a new state-of-the-art approach among probabilistic methods for image retrieval.
    \item We release PyTorch implementation of all considered methods\footnote{\url{https://github.com/tinkoff-ai/probabilistic-embeddings}}. To the best of our knowledge, it is the first public implementation of vMF-FL \cite{hasnat2017mises} and vMF-loss \cite{scott2021vmfloss} approaches.
\end{enumerate}

\section{Related Work}

\subsection{Deep Metric Learning}
The goal of metric learning is to construct a mapping from input data into latent space so that the embeddings of similar items are close to each other while those of unrelated items are far away \cite{bengio2013representation, musgrave2020metric}. Such representations can be re-used in different machine learning tasks, including verification and retrieval. In {\it verification} tasks, the algorithm solves a binary classification problem of deciding whether two items are similar or not \cite{huang2014lfw}. As distance in metric learning reflects the items' dissimilarity, classification can be performed by simple distance thresholding \cite{hadsell2006contrastive, liu2017sphereface}. In {\it image retrieval} tasks, the algorithm has to find a gallery image which is the most similar to the query image \cite{phillips2003frvt, oh2016sop}. The metric learning model estimates the similarity score between the query and each gallery image, and selects the gallery image with the maximum score.

There are two main groups of metric learning methods. The first group implements distance-based losses computed directly in an embedding space \cite{hadsell2006contrastive, weinberger2009triplet, wang2017angular, sohn2016npair, yu2019tuplet, wang2019multisimilarity}. Another group of methods solves a classification problem in such a way that representations before the last linear layer can be compared using L2 or cosine scoring \cite{wang2017normface, liu2017sphereface, wang2018cosface, wang2018amsoftmax, deng2019arcface}. In addition to the above, there are hybrid methods that have trainable target class centroids, but use distance-based losses \cite{movshovitz2017proxynca, aziere2019hardproxy, kim2020proxyanchor}.

Comprehensive comparisons of deterministic methods were recently provided by different authors \cite{musgrave2020metric, roth2020revisiting}.
In our work, we compare probabilistic methods with state-of-the-art deterministic ones \cite{movshovitz2017proxynca,wang2018cosface,wang2018amsoftmax,deng2019arcface,wang2019multisimilarity,kim2020proxyanchor}. We focus on loss functions and omit sophisticated methods such as ensembles and distillation \cite{zheng2021ensembleml,roth2021simultaneous}.

\subsection{Probabilistic Embeddings}
\begin{figure}[t]
    \vspace{.3in}
    \centering
    \includegraphics[width=\columnwidth]{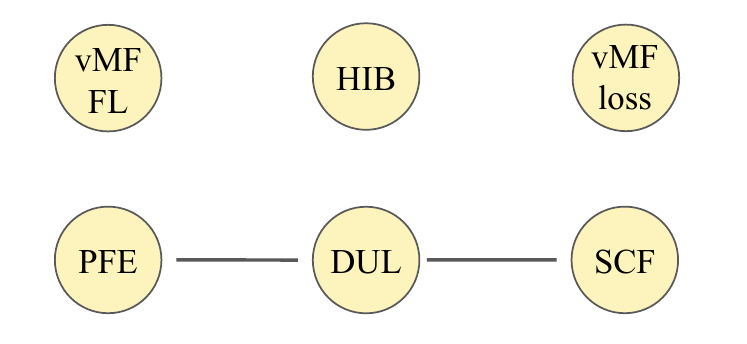}
    \caption{Probabilistic methods comparisons presented in literature. In total, 13 out of 15 comparisons are missing.}
    \label{fig:comparisons}
\end{figure}

\begin{figure*}[t]
\centering
\includegraphics[width=\textwidth]{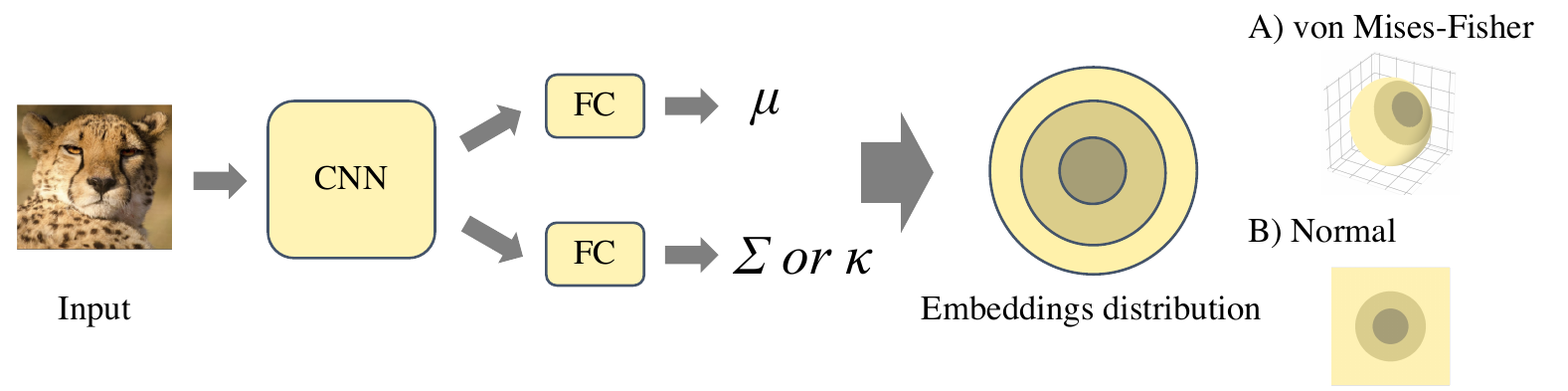}
\caption{Probabilistic embeddings prediction pipeline. CNN model is followed by mean and uncertainty prediction modules. In the case of normal distribution, $\mu$ and $\Sigma$ are predicted. For the spherical von Mises-Fisher distribution, the model predicts the mean direction $\mu$ and concentration $\kappa$.}
\label{fig:pipeline}
\end{figure*}

The ability of neural networks to predict distributions was originally studied in the work called Mixture Density Networks (MDN) \cite{bishop1994mixture}. MDN were applied to multi-dimensional regression problems with target vectors known during training. This approach is different from metric learning, where only class labels or annotated pairs are available. Distributions of latent vectors appeared in Variational Auto-Encoders (VAE) \cite{kingma2014vae}. In contrast to metric learning, the goal of VAE is high-quality data generation.

The first work that applied probabilistic embeddings to image retrieval and verification tasks is Hedged Instance Embeddings (HIB) \cite{oh2018hib}. HIB maps each input image to a normal distribution, which is parameterized by the mean vector and covariance matrix. A sampling-based training objective and matching function were proposed to train and evaluate the model. HIB was applied to the N-digit MNIST dataset of handwritten digits. It was shown that trained uncertainty improves classification and retrieval on corrupted data. Stochastic Prototype Embeddings (SPE) applied ideas from HIB to few-shot classification \cite{scott2019spe}. While SPE applies ideas from probabilistic embeddings, it did not provide a scoring function for retrieval and verification.

Independently of HIB and SPE, Probabilistic Face Embeddings (PFE) \cite{shi2019probabilistic} were proposed for face verification. In PFE, each image is mapped to a normal distribution with a mean vector produced by a pre-trained deterministic model and covariance matrix predicted by a specially trained subnetwork. PFE also introduced the Mutual Likelihood Score (MLS) for measuring similarity between distributions. Since MLS can be expressed in closed-form, PFE does not require sampling the way HIB does.

PFE gave rise to other probabilistic face verification approaches. Data Uncertainty Learning (DUL) \cite{chang2020dul} introduced end-to-end classification and two-stage regression methods called DUL-cls and DUL-reg respectively. Both methods demonstrated improvements over deterministic baselines. DUL handles uncertainty during training, but computes cosine similarity between distribution means during inference, similar to traditional metric learning approaches.

Some works combined directional statistics with probabilistic embeddings. Sphere Confidence Face (SCF) \cite{li2021spherical} is similar to DUL-reg, but applies the von Mises-Fisher distribution to model embeddings in the n-sphere latent space. Another method, called vMF-FL \cite{hasnat2017mises}, follows the same approach while applying softmax activation and training the model end-to-end. In vMF-loss \cite{scott2021vmfloss}, both image embeddings and classes centroids are modeled using the von Mises-Fisher distribution.

When it comes to evaluations, PFE and its successors \cite{ chang2020dul, li2021spherical} were not compared to HIB and vMF-FL. SCF does not refer to DUL \cite{li2021spherical}, and vMF-loss was not compared to other probabilistic methods \cite{scott2021vmfloss}. Previous studies can be found summarized in Figure \ref{fig:comparisons}. In this work, we perform missing comparisons following the best practices for metric learning evaluation \cite{musgrave2020metric}.

\subsection{Evaluation}
\label{sec:related-eval}
Metric learning and probabilistic embeddings are usually evaluated using retrieval and verification benchmarks \cite{huang2014lfw, musgrave2020metric}. In retrieval, popular metrics include Recall@1, MAP@R, Top-K accuracy, and clustering-based approaches \cite{musgrave2020metric}. Verification metrics include TPR@FPR and accuracy \cite{shi2019probabilistic, deng2019arcface}.

Recent works highlighted multiple evaluation flaws in popular metric learning benchmarks \cite{musgrave2020metric, roth2020revisiting}. Common issues include overfitting to a test set, single random seed evaluation, and unclear hyperparameter selection protocols. All the flaws highlighted in these papers are not unique to metric learning approaches, but also highly relevant to probabilistic embeddings. We address these issues by implementing an evaluation protocol inspired by Reality Check \cite{musgrave2020metric}.

\section{Background}

In this section, we review concepts related to probabilistic embeddings.

\subsection{Metric Learning}
Suppose we have a dataset of images $x_i, i=\overline{1,N}$ with class labels $y_i$. The goal of metric learning is to construct a feature mapping $z_i = f(x_i)$ where new features are close for elements of the same class and far away for elements of different classes. Ideally, there is a distance threshold $\alpha$ such that
\begin{equation}
    \begin{cases}
      \|z_1 - z_2\|_2 \le \alpha &\mbox{if } y_1 = y_2, \\
      \|z_1 - z_2\|_2 > \alpha &\mbox{if } y_1 \ne y_2.
    \end{cases}
\end{equation}
In practice, however, perfect separation is not possible, and existing methods aim to reduce the number of possible errors in different ways.

The popular {\it contrastive loss} \cite{hadsell2006contrastive} has the following form:
\begin{equation}
    \mathcal{L} = [d_p]_+ - [m - d_n]_+,
    \label{eq:contrastive}
\end{equation}
where $d_p$ is the distance between elements of the same class, $d_n$ is the distance between elements of different classes, and $m$ is a positive margin, that controls embeddings separability. This approach gave rise to many more methods \cite{weinberger2009triplet, wang2019multisimilarity, musgrave2020metric}, detailed discussion of which goes beyond the scope of our work.

Another way to choose the mapping $f$ is based on classification. Suppose $g$ is a classification model and class probabilities are computed using the following formula:
\begin{equation}
    \mathrm{P}(y = c \vert x) = \mathrm{Softmax}(g(f(x)))_c.
\end{equation}
If the model $g$ is simple enough (for example, linear), then embeddings produced by mapping $f$ are suitable for metric learning as suggested in previous works \cite{liu2017sphereface,wang2018cosface,wang2018amsoftmax,deng2019arcface}. If $g(z) = Az$ and both $z$ and rows $A_c$ are normalized, than $g$ evaluates cosine similarity scores between $z$ and $A_c$ for each $c$. In this case, the rows $A_c$ can be seen as centroids of target classes. We call them {\it target embeddings} throughout the paper.

\subsection{Probabilistic Embeddings}

\begin{table*}[t]
\caption{Comparison of probabilistic embeddings approaches.}
\label{tab:overview}
\begin{center}
\begin{minipage}{\textwidth}
\begin{tabular*}{\textwidth}{@{\extracolsep{\fill}}l|ccccc@{\extracolsep{\fill}}}
\toprule
\multirow{2}{*}{Method} & \multirow{2}{*}{Distribution} & \multirow{2}{*}{Scoring} & Training & Training & Backbone \\
&                               &                          & objective & from scratch & fine-tuning \\
\midrule
HIB      & $\mathcal{N}(\mu, \Sigma)$\footnotemark[1] & Sampling + L2   & Pair-based     & \cmark & \cmark \\
PFE      & $\mathcal{N}(\mu, \sigma^2 \mathrm{I})$\footnotemark[2] & MLS                & Pair-based     & \xmark & \xmark \\
DUL-cls  & $\mathcal{N}(\mu, \sigma^2 \mathrm{I})$\footnotemark[2] & Cosine             & Classification & \cmark & \cmark \\
DUL-reg  & $\mathcal{N}(\mu, \sigma^2 \mathrm{I})$\footnotemark[2] & Cosine             & Regression     & \xmark & \cmark \\
SCF      & $vMF(\mu, \kappa)$                      & MLS                & Regression     & \xmark & \xmark \\
vMF-FL & $vMF(\mu, \kappa)$ & Cosine & Classification & \cmark & \cmark \\
vMF-loss & $vMF(\mu, \kappa)$                      & Sampling + Cosine  & Classification & \cmark & \cmark \\
\bottomrule
\end{tabular*}

{\footnotesize $~^1$ Covariance matrix is diagonal.}

{\footnotesize $~^2$ Embeddings are L2-normalized.}
\end{minipage}
\end{center}
\end{table*}

In PE, the model predicts parameters of some distribution $\mathrm{P}(z \vert x)$ in latent space rather than a single vector as shown in \autoref{fig:pipeline}. Existing methods are summarized in Table \ref{tab:overview}. The detailed description of submodules is given below.

{\bf Distributions.}
Most methods model embeddings by normal distribution or spherical von Mises-Fisher distribution. The multivariate normal distribution is parameterized by the mean vector $\mu$ and covariance matrix $\Sigma$. Normal probability density function (PDF) is defined as
\begin{equation}
    \mathcal{N}(z;\mu, \Sigma) = \frac{1}{\sqrt{\vert 2\pi\Sigma \vert}}e^{-\frac{1}{2}(z - \mu)^T\Sigma^{-1}(z - \mu)}.
\end{equation}
In practice, most methods use a single positive number $\sigma^2$ to approximate $\Sigma$ as $\sigma^2 I$.

In many state-of-the-art metric learning approaches, embeddings are normalized to have constant length \cite{wang2018cosface, deng2019arcface}. The von Mises-Fisher (vMF) distribution is designed to model normalized data and is parameterized by mean direction $\mu, \|\mu\|_2 = 1$ and concentration parameter $\kappa$. Concentration is roughly equivalent to the inverse of normal distribution's variance. VMF probability density function for $n$-dimensional vectors is defined as
\begin{equation}
    vMF_n(z; \mu, \kappa) = \frac{\kappa^{n/2-1}}{(2\pi)^{n/2}I_{n/2-1}(\kappa)}e^{\kappa\mu^Tz},
\end{equation}
where $I_v$ denotes the modified Bessel function of the first kind for order $v$.

The mean prediction branch of PE model is usually similar to a deterministic model. Variance or concentration is predicted by a special sub-network. In the case of the von Mises-Fisher distribution, concentration $\kappa$ can be encoded into the norm of mean vector $\hat{\mu}$ so that $\kappa = \|\hat{\mu}\|_2$ and $\mu = \hat{\mu} / \|\hat{\mu}\|_2$ \cite{scott2021vmfloss}.

{\bf Scoring functions.}
An essential feature of metric learning methods is the ability to compare embeddings. In the case of probabilistic embeddings, there are three main approaches to evaluate similarity. The first one uses distance between predicted means \cite{chang2020dul}:
\begin{equation}
    S_\mu(x_1, x_2) = -d(\mu(x_1), \mu(x_2)),
\end{equation}
where $d$ is either L2 or negative cosine. This approach is equivalent to metric learning and does not take into account the uncertainty predicted by the model. Some methods estimate the expected distance via sampling \cite{scott2021vmfloss, oh2018hib}:
\begin{equation}
    S_\mathrm{E}(x_1, x_2) = -\mathrm{E}_{\substack{z_1 \sim \mathrm{P}(z_1 \vert x_1)\\z_2 \sim \mathrm{P}(z_2 \vert x_2)}} d(z_1, z_2).
    \label{eq:sampling-scoring}
\end{equation}
Finally, to eliminate sampling from comparison, a special Mutual Likelihood Score (MLS) was proposed \cite{shi2019probabilistic}. In MLS, the similarity between two distributions is computed as
\begin{equation}
    S_\mathrm{MLS}(x_1, x_2) = \int\limits_z \mathrm{P}(z \vert x_1)\mathrm{P}(z \vert x_2)dz.
\end{equation}
In the case of normal and von Mises-Fisher distributions, MLS has closed form expression \cite{shi2019probabilistic, li2021spherical}.

{\bf Training objectives.}
Probabilistic methods use different training objectives to handle distributions of embeddings. In HIB, a binary classification problem is solved by a sampling-based similarity estimation:
\begin{multline}
    \mathrm{P}(y_1 = y_2 \vert x_1, x_2) \\= \mathrm{E}_{\substack{z_1 \sim \mathrm{P}(z_1 \vert x_1)\\z_2 \sim \mathrm{P}(z_2 \vert x_2)}} g(-\alpha d(z_1, z_2) + \beta),
\end{multline}
where $\alpha$ and $\beta$ are learned parameters and $g$ is a logistic function. The loss function incorporates binary cross-entropy and KL divergence to prevent variance from converging to zero:
\begin{align}
\begin{split}
    \mathcal{L}_{HIB} = &-\log \mathrm{P}(y_1 = y_2 \vert x_1, x_2) \\ &+ KL(\mathrm{P}(z_1 \vert x_1) \| \mathcal{N}(0, \mathrm{I}))\\ &+ KL(\mathrm{P}(z_2 \vert x_2) \| \mathcal{N}(0, \mathrm{I})).
\end{split}
\end{align}

Probabilistic Face Embeddings (PFE) \cite{shi2019probabilistic} take a pretrained deterministic model (like CosFace or ArcFace) for mean prediction and train a variance prediction module to maximize the MLS score between elements of the same class from the batch.

DUL-cls \cite{chang2020dul} samples from predicted distribution and computes deterministic classification loss (CosFace or ArcFace) for each sample. KL-divergence is also added to the loss, similar to HIB.
Another method, called DUL-reg \cite{chang2020dul}, takes pretrained target class centroids $z_y$ and performs regression to these centroids via cross-entropy loss, similar to Mixture Density Networks (MDN) \cite{bishop1994mixture}:
\begin{equation}
    \mathcal{L}_{Reg} = -\log \mathrm{P}(z_{y} \vert x).
\end{equation}

The SCF \cite{li2021spherical} training is close to DUL-reg, but uses von Mises-Fisher distribution instead of normal. Both SCF and DUL-reg do regression to the pretrained centroids. In contrast to these methods, vMF-FL \cite{hasnat2017mises} applies softmax activation to posteriors and trains the model end-to-end:
\begin{equation}
    \mathcal{L}_{vMF\mbox{-}FL} = -\log \frac{\mathrm{P}(z_{y} \vert x)}{\sum\limits_c \mathrm{P}(z_c \vert x)}.
\end{equation}

The vMF-loss approach \cite{scott2021vmfloss} models two distributions $\mathrm{P}(z \vert x)$ and $\mathrm{P}(z \vert y)$. The method then minimizes the upper bound of the expected softmax loss function, which takes the form:
\begin{equation}
    \mathcal{L}_{vMF} = -\mathrm{E}_{\substack{
    z\sim\mathrm{P}(z \vert x)\\
    z_1\sim\mathrm{P}(z \vert y = 1)\\
    \dots\\
    z_c\sim\mathrm{P}(z \vert y = c)}}\log\frac{e^{\beta d(z, z_y)}}{\sum_c e^{\beta d(z, z_c)}},
\end{equation}
where $d$ is cosine similarity and $\beta$ is a learned parameter.
At inference time, vMF-loss uses sampling-based cosine scoring from Equation \ref{eq:sampling-scoring}.

\section{Evaluation Protocol}
\label{sec:protocol}
To address previously highlighted evaluation flaws (see Section \ref{sec:related-eval} for details), we implemented an evaluation protocol inspired by recent metric learning benchmarks \cite{musgrave2020metric, roth2020revisiting}. 
In this section, we describe the used datasets, metrics, and training procedure.

\subsection{Datasets}
\label{sec:datasets}
We use the CUB200-2011 (CUB200) \cite{welinder2010cub}, \mbox{Cars196 \cite{krause2013cars}}, In-shop clothes (In-shop) \cite{liu2016inshop}, and Stanford Online Products (SOP) \cite{oh2016sop} datasets for evaluation. While In-shop and SOP have class-disjoint train/test splits provided by the authors, CUB200 and Cars196 are closed-set classification datasets. In order to convert them into class-disjoint retrieval benchmarks, we split classes into two equal subsets, using the first one for development and the second for testing \cite{wu2017datasetsplit}. The first quarter of development classes is used for validation and the remaining part is used for training.

We construct a verification testset for each benchmark. To do this, we sample pairs of elements with the same class (positives) and pairs of elements with different classes (negatives). Both the number of positives and the number of negatives are equal to the size of the source classification dataset.

We also build corrupted variants of the datasets for confidence prediction evaluation. To reduce the amount of information in each image, we extract a central crop with a random size between $0.5$ and $1$ of the image's original size. The size of the crop is used as the ground truth estimation of image quality.

\subsection{Metrics}
We evaluate the image retrieval quality using the Recall@1 and MAP@R metrics \cite{musgrave2020metric}. Both metrics are evaluated in embedding space rather than using classification head outputs. We also evaluate the verification accuracy as the maximum binary classification accuracy among decision thresholds on verification testsets described above.

\subsection{Model Architecture and Training}
We use a BN-inception convolutional neural network (CNN) for predicting embeddings following recent metric learning benchmarks \cite{musgrave2020metric, roth2020revisiting}. We freeze batch normalization layers during training and replace the average pooling with a sum of average and max pooling layers as suggested by previous works \cite{kim2020proxyanchor, jun2019multihead}. The embedding size is fixed to 128 in all experiments. In probabilistic methods, an extra sub-network is connected to the CNN output, as shown in Figure \ref{fig:pipeline}. We use a single fully-connected layer for mean prediction and 3 fully-connected layers with ReLU activations for confidence estimation following SCF \cite{li2021spherical}.

We use different learning rates for the CNN and final classification layer following \cite{kim2020proxyanchor, scott2021vmfloss}. Both learning rates are considered model hyperparameters. We use stochastic gradient descent with momentum 0.9 and weight decay $0.0001$ for optimization. Training stops when validation quality does not improve for 10 epochs. The model from the epoch with the best validation quality (MAP@R) is used for testing.

During training, we augment data using random horizontal flip, brightness, contrast, and saturation jittering with factors between $0.75$ and $1.25$, random center crop with a scale between $0.16$ and $1$ and aspect ratio between $0.75$ and $1.33$. The cropped image is resized to 224 pixels on each side. During testing, we resize the image to 256 pixels on the shortest side and center-crop image to a size 224.

To tune the methods' hyperparameters, we run 50 iterations of Bayesian Hyperparameter Search. The set of hyperparameters with maximum validation MAP@R is evaluated using 5 random seeds. We report the mean and STD of the produced set of metrics.

\section{Experiments}
\subsection{Probabilistic Embeddings Performance}
\label{sec:exp-general}




\begin{table*}[tp]
\caption{Comparison of deterministic and probabilistic methods on Cars196 and CUB200 datasets. Deterministic methods are at the top and probabilistic in the middle of the table. Results for our non-standard variant of the probabilistic DUL method are presented at the bottom. Results greater than those of deterministic methods (ArcFace -- Proxy-NCA) are underlined. Methods with maximum performance are shown in bold. All values are reported in percentages.}
\label{tab:general-small}
\begin{center}
\begin{minipage}{\textwidth}
\begin{tabular*}{\textwidth}{@{\extracolsep{\fill}}l|ccc|ccc@{\extracolsep{\fill}}}
\toprule
\multirow{2}{*}{Method} & \multicolumn{3}{c|}{Cars196} & \multicolumn{3}{c}{CUB200} \\
& Recall@1 & MAP@R & Accuracy & Recall@1 & MAP@R & Accuracy \\
\midrule
ArcFace       & 71.3 $\pm$ 0.4 & 18.4 $\pm$ 0.1 & 82.1 $\pm$ 0.3 & \bf 61.0 $\pm$ 0.5 & \bf 22.3 $\pm$ 0.4 & 84.1 $\pm$ 0.2 \\
CosFace       & 72.7 $\pm$ 0.5 & 19.0 $\pm$ 0.2 & 80.4 $\pm$ 0.2 & 57.1 $\pm$ 1.0 & 20.0 $\pm$ 0.8 & 83.9 $\pm$ 0.4 \\
Proxy-Anc. & \bf 75.7 $\pm$ 0.6 & 19.2 $\pm$ 0.5 & 79.0 $\pm$ 0.3 & 60.0 $\pm$ 0.9 & 21.2 $\pm$ 0.3 & 83.3 $\pm$ 0.3 \\
Multi-sim. & 72.0 $\pm$ 0.5 & 18.3 $\pm$ 0.3 & 81.2 $\pm$ 0.3 & 58.1 $\pm$ 0.4 & 20.7 $\pm$ 0.1 & 84.7 $\pm$ 0.2 \\
Proxy-NCA & 63.3 $\pm$ 0.7 & 15.1 $\pm$ 0.6 & 81.3 $\pm$ 0.2 & 49.3 $\pm$ 0.4 & 15.0 $\pm$ 0.2 & 82.2 $\pm$ 0.3 \\
\midrule
HIB           & 39.6 $\pm$ 0.9 & 6.7 $\pm$ 0.3 & 80.4 $\pm$ 0.2 & 43.3 $\pm$ 0.8 & 11.8 $\pm$ 0.3 & 81.4 $\pm$ 0.2 \\
PFE           & 71.0 $\pm$ 0.3 & 18.0 $\pm$ 0.1 & \underline{82.2 $\pm$ 0.2} & 60.8 $\pm$ 0.4 & \bf \underline{22.3 $\pm$ 0.4} & \bf \underline{85.1 $\pm$ 0.2} \\
DUL-cls       & 73.6 $\pm$ 0.7 & \underline{19.5 $\pm$ 0.2} & 81.3 $\pm$ 0.1 & 59.8 $\pm$ 0.4 & 21.2 $\pm$ 0.4 & 83.7 $\pm$ 0.3 \\
DUL-reg       & 70.1 $\pm$ 0.7 & 19.1 $\pm$ 0.2 & \bf \underline{82.6 $\pm$ 0.2} & 58.7 $\pm$ 0.7 & 21.3 $\pm$ 0.4 & 84.3 $\pm$ 0.2 \\
SCF           & 70.4 $\pm$ 0.4 & 18.3 $\pm$ 0.1 & \underline{82.4 $\pm$ 0.2} & 57.6 $\pm$ 1.0 & 20.2 $\pm$ 0.6 & 84.0 $\pm$ 0.3 \\
vMF-FL & 63.2 $\pm$ 0.5 & 11.9 $\pm$ 0.1 & 77.7 $\pm$ 0.4 & 52.5 $\pm$ 0.6 & 15.2 $\pm$ 0.3 & 81.3 $\pm$ 1.0 \\
vMF-loss & 62.8 $\pm$ 0.3 & 14.5 $\pm$ 0.2 & 80.2 $\pm$ 0.3 & 52.7 $\pm$ 1.0 & 17.8 $\pm$ 0.7 & 83.2 $\pm$ 0.4 \\
\midrule
DUL-reg-cls   & 72.7 $\pm$ 0.4 & \bf \underline{20.1 $\pm$ 0.3} & 81.9 $\pm$ 0.4 & 60.1 $\pm$ 0.8 & 21.6 $\pm$ 0.4 & 84.5 $\pm$ 0.3 \\
\bottomrule
\end{tabular*}
\end{minipage}
\end{center}
\bigskip
\caption{Comparison of deterministic and probabilistic methods on In-shop and SOP datasets. Deterministic methods are at the top and probabilistic in the middle of the table. Results for our non-standard variant of probabilistic DUL method are presented at the bottom. Results greater than those of deterministic methods (ArcFace -- Proxy-NCA) are underlined. Methods with maximum performance are shown in bold. All values are reported in percentages.}
\label{tab:general-large}
\begin{center}
\begin{minipage}{\textwidth}
\begin{tabular*}{\textwidth}{@{\extracolsep{\fill}}l|ccc|ccc@{\extracolsep{\fill}}}
\toprule
\multirow{2}{*}{Method} & \multicolumn{3}{c|}{In-shop} & \multicolumn{3}{c}{SOP} \\
& Recall@1 & MAP@R & Accuracy & Recall@1 & MAP@R & Accuracy \\
\midrule
ArcFace       & 78.6 $\pm$ 0.3 & 37.6 $\pm$ 0.2 & 92.5 $\pm$ 0.2 & 59.7 $\pm$ 0.3 & 30.6 $\pm$ 0.2 & 89.9 $\pm$ 0.1 \\
CosFace       & 87.5 $\pm$ 0.2 & 45.1 $\pm$ 0.1 & 92.5 $\pm$ 0.1 & 67.4 $\pm$ 0.4 & 37.5 $\pm$ 0.3 & 91.0 $\pm$ 0.1 \\
Proxy-Anc. & 71.1 $\pm$ 0.2 & 31.2 $\pm$ 0.1 & 91.0 $\pm$ 0.2 & 59.8 $\pm$ 0.2 & 30.5 $\pm$ 0.1 & 90.3 $\pm$ 0.1 \\
Multi-sim. & 86.5 $\pm$ 0.4 & \bf 47.5 $\pm$ 0.4 & \bf 95.4 $\pm$ 0.1 & 66.9 $\pm$ 0.2 & 37.4 $\pm$ 0.1 & \bf 92.5 $\pm$ 0.0 \\
Proxy-NCA & 67.5 $\pm$ 0.9 & 28.7 $\pm$ 0.4 & 90.6 $\pm$ 0.3 & 49.6 $\pm$ 0.6 & 22.8 $\pm$ 0.4 & 88.4 $\pm$ 0.2 \\
\midrule
HIB           & 52.6 $\pm$ 1.9 & 21.9 $\pm$ 1.1 & 92.0 $\pm$ 0.3 & 55.3 $\pm$ 0.5 & 27.4 $\pm$ 0.4 & 91.2 $\pm$ 0.1 \\
PFE           & 87.2 $\pm$ 0.2 & 44.5 $\pm$ 0.1 & 89.6 $\pm$ 0.1 & 67.0 $\pm$ 0.4 & 37.1 $\pm$ 0.3 & 90.4 $\pm$ 0.1 \\
DUL-cls       &  \underline {88.9 $\pm$ 0.1} & 47.0 $\pm$ 0.1 & 93.8 $\pm$ 0.1 & \underline {68.8 $\pm$ 0.3} & \underline {39.4 $\pm$ 0.3} & 91.7 $\pm$ 0.1 \\
DUL-reg       & \underline {88.4 $\pm$ 0.1} & 46.4 $\pm$ 0.2 & 93.6 $\pm$ 0.1 & \underline {68.3 $\pm$ 0.3} & \underline {38.7 $\pm$ 0.3} & 91.9 $\pm$ 0.1 \\
SCF           & 87.3 $\pm$ 0.2 & 44.7 $\pm$ 0.1 & 89.5 $\pm$ 0.2 & \underline {67.4 $\pm$ 0.4} & \underline {37.5 $\pm$ 0.3} & 91.2 $\pm$ 0.1 \\
vMF-FL & 40.1 $\pm$ 0.5 & 13.8 $\pm$ 0.3 & 82.9 $\pm$ 0.3 & 46.1 $\pm$ 0.5 & 20.2 $\pm$ 0.3 & 81.9 $\pm$ 0.4 \\
vMF-loss & 61.5 $\pm$ 0.7 & 24.7 $\pm$ 0.4 & 80.0 $\pm$ 0.5 & 43.0 $\pm$ 0.5 & 18.8 $\pm$ 0.2 & 81.1 $\pm$ 1.0 \\
\midrule
DUL-reg-cls    & \bf \underline{89.0 $\pm$ 0.1} & 47.2 $\pm$ 0.1 & 94.3 $\pm$ 0.0 & \bf \underline{69.2 $\pm$ 0.3} & \bf \underline{39.9 $\pm$ 0.3} & 92.3 $\pm$ 0.2 \\
\bottomrule
\end{tabular*}
\end{minipage}
\end{center}
\vspace{.5in}
\end{table*}

As some methods were previously applied only to the face verification task, and many methods were not directly compared, we asked the following questions: (1) do probabilistic embeddings improve verification and retrieval quality compared to deterministic metric learning baselines and (2) which methods are more suited for the considered tasks?

We compared different probabilistic methods with ArcFace \cite{deng2019arcface}, CosFace \cite{wang2018cosface}, Proxy-Anchor \cite{kim2020proxyanchor}, Multi-similarity \cite{wang2019multisimilarity}, and Proxy-NCA \cite{movshovitz2017proxynca}.
Results are reported in Table \ref{tab:general-small} and Table \ref{tab:general-large}.

To summarize the results, probabilistic methods achieve higher quality than deterministic baselines in 7 out of 12 comparisons (3 metrics for 4 datasets). On small datasets (Cars196 and CUB200), results are mixed. On In-shop and SOP multi-similarity achieves the highest verification accuracy and DUL methods are better in 3 out of 4 retrieval comparisons. At the same time DUL outperforms its backbone Cosface model in both verification and retrieval.

We measured the training time of considered methods on a single Nvidia Tesla V100 GPU. According to our results, probabilistic methods add at most 3\% computation overhead compared to ArcFace and CosFace, except for the vMF-loss approach. The latter's training time depends on the number of target classes, and ranges from 6\% (1.71 ms vs 1.62 ms per sample) on Cars196 to 30\% (2.07 ms vs 1.62 ms per sample) on the SOP dataset in comparison to ArcFace and CosFace.

\subsection{Target Choice}
Some probabilistic methods, such as SCF and DUL-reg, train regression to precomputed target embeddings. In contrast, DUL-cls trains a model from scratch, including target embeddings. We implemented a new variant of the DUL method, which applies DUL-reg to the target embeddings, pretrained by the DUL-cls method, instead of ArcFace or CosFace. Evaluation results of this method, called DUL-reg-cls, are presented at the bottom of Table \ref{tab:general-small} and Table \ref{tab:general-large}.

DUL-reg-cls improves DUL performance in 10 out of 12 benchmarks, and achieves top-performing results in 4 out of 12 cases. On In-shop and SOP datasets DUL-reg-cls achieves state-of-the-art performance among probabilistic methods and outperforms deterministic approaches in terms of retrieval quality in 3 out of 4 comparisons. We conclude there, that the choice of target embeddings largely affects probabilistic embeddings performance. Many metric learning approaches, such as classification-based losses and proxy-anchors, provide target embeddings. For these methods, probabilistic regression can be used to fine-tune the model after training.
\begin{table*}[tp]
\caption{Comparison of the vMF and normal distributions for probabilistic methods in terms of MAP@R (\%). Distributions used in the source papers are underlined.}
\label{tab:vmf-normal}
\begin{center}
\begin{minipage}{\textwidth}
\begin{tabular*}{\textwidth}{@{\extracolsep{\fill}}lc|cccc@{\extracolsep{\fill}}}
\toprule
Method & Distribution & Cars196 & Cub200 & In-shop & SOP \\
\midrule
\multirow{2}{*}{PFE}         & \underline{Normal}    & 18.0 $\pm$ 0.1 & 22.3 $\pm$ 0.4 & 44.5 $\pm$ 0.1 & 37.1 $\pm$ 0.3 \\
     & vMF       & 17.5 $\pm$ 0.2 & 22.3 $\pm$ 0.4 & 43.9 $\pm$ 0.2 & 36.9 $\pm$ 0.3 \\
\midrule
\multirow{2}{*}{DUL-cls}     & \underline{Normal}    & 19.5 $\pm$ 0.2 & 21.2 $\pm$ 0.4 & 47.0 $\pm$ 0.1 & 39.4 $\pm$ 0.3 \\
 & vMF       & 4.0 $\pm$ 1.8 & 0.9 $\pm$ 0.5 & 4.1 $\pm$ 0.3 & 12.4 $\pm$ 0.6 \\
\midrule
\multirow{2}{*}{DUL-reg}     & \underline{Normal}    & 19.1 $\pm$ 0.2 & 21.3 $\pm$ 0.4 & 46.4 $\pm$ 0.2 & 38.7 $\pm$ 0.3 \\
 & vMF       & 19.0 $\pm$ 0.3 & 22.1 $\pm$ 0.1 & 46.4 $\pm$ 0.2 & 38.6 $\pm$ 0.3 \\
\midrule
\multirow{2}{*}{SCF}  & Normal    & 18.3 $\pm$ 0.1 & 20.4 $\pm$ 0.2 & 45.0 $\pm$ 0.1 & 37.5 $\pm$ 0.3 \\
        & \underline{vMF}       & 18.3 $\pm$ 0.1 & 20.2 $\pm$ 0.6 & 44.7 $\pm$ 0.1 & 37.5 $\pm$ 0.3 \\
\midrule
\multirow{2}{*}{vMF-FL}  & Normal    & 11.9 $\pm$ 0.5 & 15.5 $\pm$ 0.3 & 13.7 $\pm$ 0.3 & 20.3 $\pm$ 0.3  \\
        & \underline{vMF}       & 11.9 $\pm$ 0.1 & 15.2 $\pm$ 0.3 & 13.8 $\pm$ 0.3 & 20.2 $\pm$ 0.3 \\
\bottomrule
\end{tabular*}
\end{minipage}
\end{center}

\bigskip

\caption{Comparison of scoring functions in terms of MAP@R (\%). Scoring functions used in the original papers are underlined.}
\label{tab:scoring}
\begin{center}
\begin{minipage}{\textwidth}
\begin{tabular*}{\textwidth}{@{\extracolsep{\fill}}lc|cccc@{\extracolsep{\fill}}}
\toprule
Method & Scoring & Cars196 & Cub200 & In-shop & SOP \\
\midrule
\multirow{3}{*}{HIB}  & L2 & 6.7 $\pm$ 0.3 & 11.8 $\pm$ 0.3 & 21.9 $\pm$ 1.1 & 27.4 $\pm$ 0.4 \\
& \underline{Sampling + L2} & 6.7 $\pm$ 0.3  & 11.8 $\pm$ 0.3 & 21.9 $\pm$ 1.1 & 27.4 $\pm$ 0.4 \\
& MLS       & 6.6 $\pm$ 0.2 & 11.3 $\pm$ 0.2 & 21.7 $\pm$ 1.1 & 26.0 $\pm$ 0.2 \\
\midrule
\multirow{2}{*}{PFE}      & Sampling + Cosine & 17.8 $\pm$ 0.1 & 20.7 $\pm$ 0.5 & 44.7 $\pm$ 0.1 & 36.7 $\pm$ 0.2 \\
& \underline{MLS}       & 18.0 $\pm$ 0.1 & 22.3 $\pm$ 0.4 & 44.5 $\pm$ 0.1 & 37.1 $\pm$ 0.3 \\
\midrule
\multirow{3}{*}{DUL-cls}     & \underline{Cosine}    & 19.5 $\pm$ 0.2 & 21.2 $\pm$ 0.4 & 47.0 $\pm$ 0.1 & 39.4 $\pm$ 0.3 \\
& Sampling + Cosine & 19.5 $\pm$ 0.2 & 21.1 $\pm$ 0.4 & 47.0 $\pm$ 0.1 & 39.4 $\pm$ 0.3 \\
& MLS       & 18.4 $\pm$ 0.2 & 20.4 $\pm$ 0.4 & 46.8 $\pm$ 0.1 & 37.9 $\pm$ 0.3 \\
\midrule
\multirow{3}{*}{DUL-reg}     & \underline{Cosine}    & 19.1 $\pm$ 0.2 & 21.3 $\pm$ 0.4 & 46.4 $\pm$ 0.2 & 38.7 $\pm$ 0.3 \\
& Sampling + Cosine & 18.3 $\pm$ 0.3 & 18.4 $\pm$ 0.5 & 45.7 $\pm$ 0.2 & 37.5 $\pm$ 0.2 \\
& MLS       & 19.0 $\pm$ 0.2 & 20.1 $\pm$ 0.5 & 46.1 $\pm$ 0.2 & 38.7 $\pm$ 0.3 \\
\midrule
\multirow{2}{*}{SCF}      & Sampling + Cosine & 16.2 $\pm$ 0.1 & 17.7 $\pm$ 0.5 & 43.3 $\pm$ 0.2 & 34.5 $\pm$ 0.3 \\
& \underline{MLS}       & 18.3 $\pm$ 0.1 & 20.2 $\pm$ 0.6 & 44.7 $\pm$ 0.1 & 37.5 $\pm$ 0.3 \\
\midrule
\multirow{3}{*}{vMF-FL}     &  \underline{Cosine} & 11.9 $\pm$ 0.1 & 15.2 $\pm$ 0.3 & 13.8 $\pm$ 0.3 & 20.2 $\pm$ 0.3 \\
& Sampling + Cosine & 6.4 $\pm$ 1.7 &  7.7 $\pm$ 0.3 & 4.4 $\pm$ 0.4 & 7.9 $\pm$ 0.3 \\
& MLS       & 7.3 $\pm$ 1.6 & 10.3 $\pm$ 0.3 & 11.7 $\pm$ 0.4 & 16.1 $\pm$ 0.4 \\
\midrule
\multirow{3}{*}{vMF-loss}     & Cosine    & 17.0 $\pm$ 0.1 & 19.8 $\pm$ 0.7 & 25.0 $\pm$ 0.4 & 19.4 $\pm$ 0.2 \\
& \underline{Sampling + Cosine} & 14.5 $\pm$ 0.2 & 17.8 $\pm$ 0.7 & 24.7 $\pm$ 0.4 & 18.8 $\pm$ 0.2 \\
& MLS       & 15.1 $\pm$ 0.5 & 17.9 $\pm$ 0.4 & 21.2 $\pm$ 0.2 & 19.2 $\pm$ 0.3 \\
\bottomrule
\end{tabular*}
\end{minipage}
\end{center}
\vspace{.5in}
\end{table*}
\begin{table*}[tp]
\caption{Recall@1 (\%) versus confidence-based error detection accuracy (CEDA, \%) on the truncated training part of the dataset.}
\label{tab:confidence-generalization-train}
\begin{center}
\begin{minipage}{\textwidth}
\begin{tabular*}{\textwidth}{@{\extracolsep{\fill}}l|cc|cc|cc|cc@{\extracolsep{\fill}}}
\toprule
\multirow{2}{*}{Method} & \multicolumn{2}{c|}{Cars196} & \multicolumn{2}{c|}{CUB200} & \multicolumn{2}{c|}{In-shop} & \multicolumn{2}{c}{SOP} \\
& Recall@1 & CEDA & Recall@1 & CEDA & Recall@1 & CEDA & Recall@1 & CEDA \\
\midrule
Baseline      & 89.9 & 89.9 & 94.1 & 94.1 & 90.6 & 90.6 & 83.4 & 83.4 \\
\midrule
HIB           & 62.7 & 63.0 & 86.6 & 86.9 & 59.8 & 62.2 & 67.4 & 67.4 \\
PFE           & 89.9 & 91.0 & 94.1 & 94.6 & 90.6 & 91.3 & 83.4 & 83.4 \\
DUL-cls       & 90.8 & 91.6 & 93.1 & 93.8 & 91.1 & 91.9 & 83.4 & 83.5 \\
DUL-reg       & 88.0 & 89.5 & 92.4 & 93.1 & 90.7 & 91.7 & 83.7 & 83.8 \\
SCF           & 89.9 & 91.2 & 93.8 & 94.4 & 90.6 & 91.3 & 83.4 & 83.4 \\
vMF-FL & 96.6 & 96.8 & 97.4 & 97.5 & 99.7 & 99.8 & 99.5 & 99.5 \\
vMF-loss      & 89.6 & 90.5 & 93.0 & 93.7 & 79.4 & 80.8 & 63.7 & 64.6 \\
\midrule
DUL-reg-cls   & 89.8 & 91.2 & 93.3 & 94.0 & 91.1 & 92.0 & 83.2 & 83.4 \\
\bottomrule
\end{tabular*}
\end{minipage}
\end{center}
\end{table*}

\begin{table*}[tp]
\caption{Recall@1 (\%) versus confidence-based error detection accuracy (CEDA, \%) on the testing part of the dataset.}
\label{tab:confidence-generalization-test}
\begin{center}
\begin{minipage}{\textwidth}
\begin{tabular*}{\textwidth}{@{\extracolsep{\fill}}l|cc|cc|cc|cc@{\extracolsep{\fill}}}
\toprule
\multirow{2}{*}{Method} & \multicolumn{2}{c|}{Cars196} & \multicolumn{2}{c|}{CUB200} & \multicolumn{2}{c|}{In-shop} & \multicolumn{2}{c}{SOP} \\
& Recall@1 & CEDA & Recall@1 & CEDA & Recall@1 & CEDA & Recall@1 & CEDA \\
\midrule
Baseline      & 71.3 & 71.3 & 61.0 & 61.0 & 87.5 & 87. & 67.4 & 67.4 \\
\midrule
HIB           & 39.6.0 & 60.2 & 43.3 & 57.0 & 52.6 & 57.1 & 55.3 & 55.6 \\
PFE           & 71.0 & 71.8 & 60.8 & 63.6 & 87.2 & 87.6 & 67.0 & 69.0 \\
DUL-cls       & 73.6 & 73.4 & 59.8 & 60.5 & 88.9 & 88.9 & 68.8 & 69.5 \\
DUL-reg       & 70.1 & 70.9 & 58.7 & 59.8 & 88.4 & 88.5 & 68.3 & 69.4 \\
SCF           & 70.4 & 71.8 & 57.6 & 59.9 & 87.3 & 87.6 & 67.4 & 68.6 \\
vMF-FL & 63.2 & 63.5 & 52.5 & 54.6 & 40.1 & 60.1 & 46.1 & 53.9 \\
vMF-loss      & 70.7 & 71.5 & 57.3 & 58.6 & 62.1 & 65.8 & 43.5 & 57.5 \\
\midrule
DUL-reg-cls   & 72.7 & 73.4 & 60.1 & 61.6 & 89.0 & 89.0 & 69.2 & 70.6 \\
\bottomrule
\end{tabular*}
\end{minipage}
\end{center}
\end{table*}

\begin{table*}[tp]
\caption{Spearman's rank correlation coefficient between confidences and image quality on corrupted datasets. We used maximum class probability predicted by a deterministic model (ArcFace for Cars196 and CUB200, and CosFace for In-shop and SOP) as a baseline.}
\label{tab:lossy-scc}
\begin{center}
\begin{small}
\begin{sc}
\begin{tabular}{l|cccc}
\toprule
Method & Cars196 & CUB200 & In-shop & SOP \\
\midrule
Baseline-prob     & 0.32 $\pm$ 0.02 & 0.06 $\pm$ 0.01 & 0.14 $\pm$ 0.01 & 0.05 $\pm$ 0.00 \\
Baseline-norm     & 0.50 $\pm$ 0.02 & 0.54 $\pm$ 0.02 & 0.23 $\pm$ 0.01 & 0.43 $\pm$ 0.02 \\
\midrule
HIB         & 0.27 $\pm$ 0.10 & 0.22 $\pm$ 0.12 & 0.04 $\pm$ 0.08 & 0.20 $\pm$ 0.03 \\
PFE         & 0.60 $\pm$ 0.06 & 0.51 $\pm$ 0.03 & 0.16 $\pm$ 0.02 & 0.41 $\pm$ 0.02 \\
DUL-cls     & 0.60 $\pm$ 0.02 & 0.56 $\pm$ 0.05 & 0.22 $\pm$ 0.00 & 0.43 $\pm$ 0.02  \\
DUL-reg     & 0.65 $\pm$ 0.02 & 0.51 $\pm$ 0.02 & 0.53 $\pm$ 0.02 & \bf 0.66 $\pm$ 0.07 \\
SCF         & 0.66 $\pm$ 0.03 & 0.46 $\pm$ 0.03 & 0.20 $\pm$ 0.04 & 0.50 $\pm$ 0.02 \\
vMF-FL & \bf 0.72 $\pm$ 0.02 & \bf 0.65 $\pm$ 0.01 & \bf 0.64 $\pm$ 0.04 & 0.28 $\pm$ 0.04 \\
vMF-loss & 0.50 $\pm$ 0.02  & 0.55 $\pm$ 0.05 & 0.17 $\pm$ 0.01 & 0.33 $\pm$ 0.03 \\
\midrule
DUL-reg-cls & 0.67 $\pm$ 0.01 & 0.56 $\pm$ 0.05 & 0.50 $\pm$ 0.04 & 0.62 $\pm$ 0.02 \\
\bottomrule
\end{tabular}
\end{sc}
\end{small}
\end{center}
\end{table*}

\begin{figure*}[t]
    \centering
    \includegraphics[width=\linewidth]{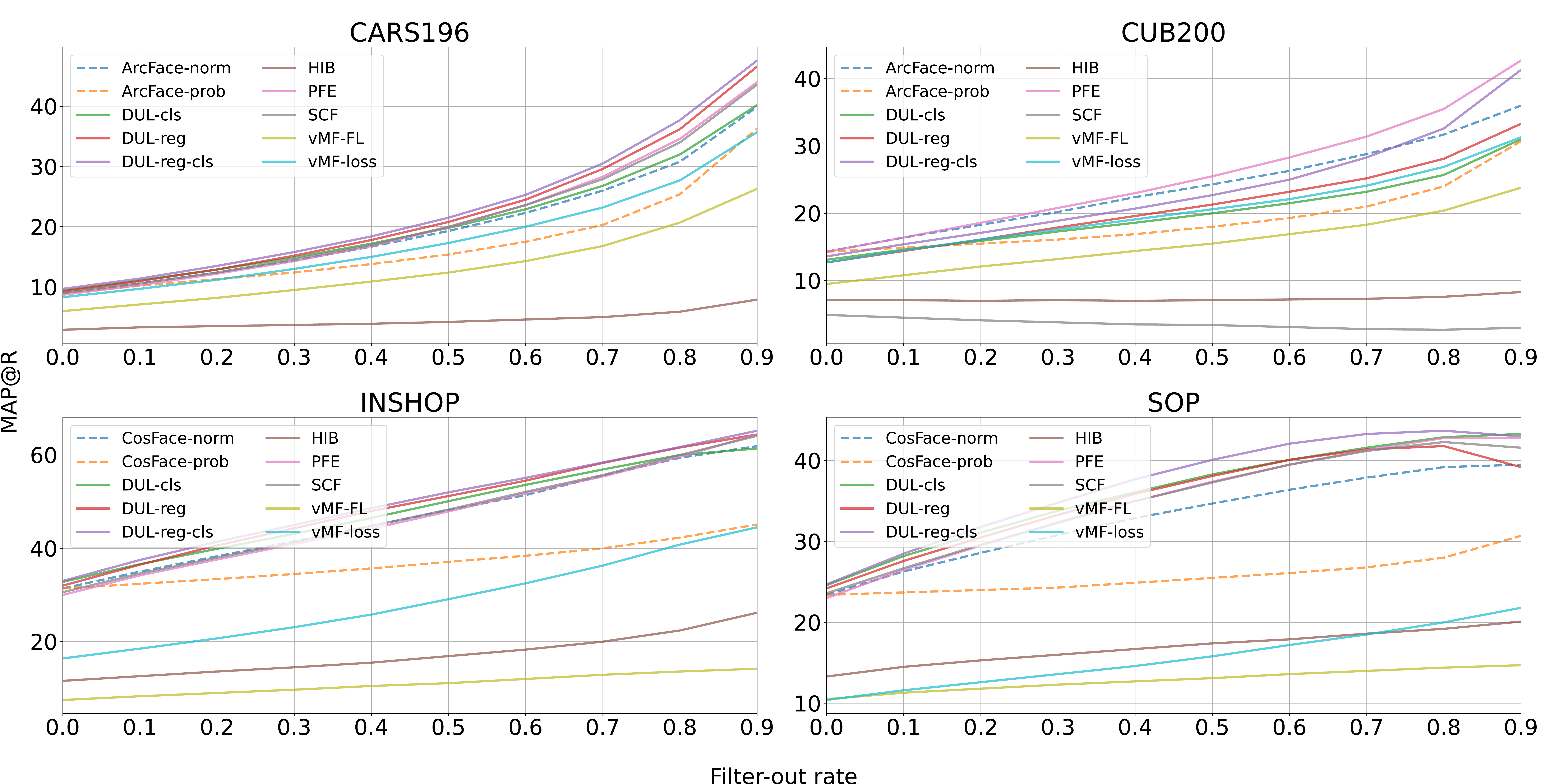}
    \caption{MAP@R for different filter-out rates on corrupted datasets. Probabilistic methods assign confidence to each input image. Confidence is not available for deterministic methods ArcFace and CosFace. For them we use maximum classification probability.}
    \label{filter-out}
\end{figure*}

\subsection{Distribution Type}
As shown in Table \ref{tab:overview}, some probabilistic methods use the normal distribution to model embeddings, and the others are based on the von Mises-Fisher distribution. While the vMF distribution was specially designed for spherical spaces \cite{li2021spherical, scott2021vmfloss}, a normal distribution is more computationally effective when dealing with sampling or re-parametrization trick \cite{kingma2014vae, davidson2018vmfreparam}. Furthermore, many methods apply the normal distribution to normalized embeddings, making it similar to vMF. In most methods, one distribution can be used in place of another. Finding the one which is the most preferable in practice is the main question of this experiment.

We compared variants of probabilistic methods based on the vMF and normal distributions. Results are presented in Table \ref{tab:vmf-normal}. HIB is excluded from this experiment, as it uses unnormalized embeddings which cannot be modeled via the von Mises-Fisher distribution \cite{li2021spherical}. The vMF-loss method was not compared with its normal-based counterpart as its special loss function \cite{scott2021vmfloss} requires sophisticated derivation, which goes beyond the scope of this work. In most cases, there is little difference between methods based on different types of distributions. One exception is the DUL-cls method, as its quality is largely reduced with the vMF distribution.

\subsection{Scoring Functions}
The scoring function is an essential part of the probabilistic embeddings pipeline, which is usually independent of the training objective. While probabilistic methods discussed in this work are based on different scoring approaches, the effect of scoring was not studied in the original papers. Most scoring functions are interchangeable, and it is important to determine which one is better to use in practice.

According to Table \ref{tab:overview}, there are three groups of probabilistic methods. PFE and SCF use Mutual Likelihood Scoring (MLS), HIB and vMF-loss apply deterministic scoring after sampling from predicted distribution, and DUL performs simple cosine scoring between distribution means. As PFE and SCF train only variance prediction module, cosine scoring between distribution means is equivalent to the corresponding deterministic model (ArcFace or CosFace). We thus exclude these combinations from the comparison.

Keeping this in mind, we compared probabilistic methods with different scoring functions. Results are presented in Table \ref{tab:scoring}. In most cases, the distance between distribution means provides on-par or better results than other scoring functions.
For PFE and SCF cosine scoring is meaningless, as was mentioned above. For these methods MLS usually achieves higher quality than the sampling-based approach.

\subsection{Confidence and Data Quality Estimation}

\begin{figure*}
  \centering
  \begin{subfigure}{0.45\linewidth}
    \begin{minipage}{0.05\linewidth}
      \rotatebox{90}{Cars196}
    \end{minipage}
    \begin{minipage}{0.9\linewidth}
        \includegraphics[width=\linewidth]{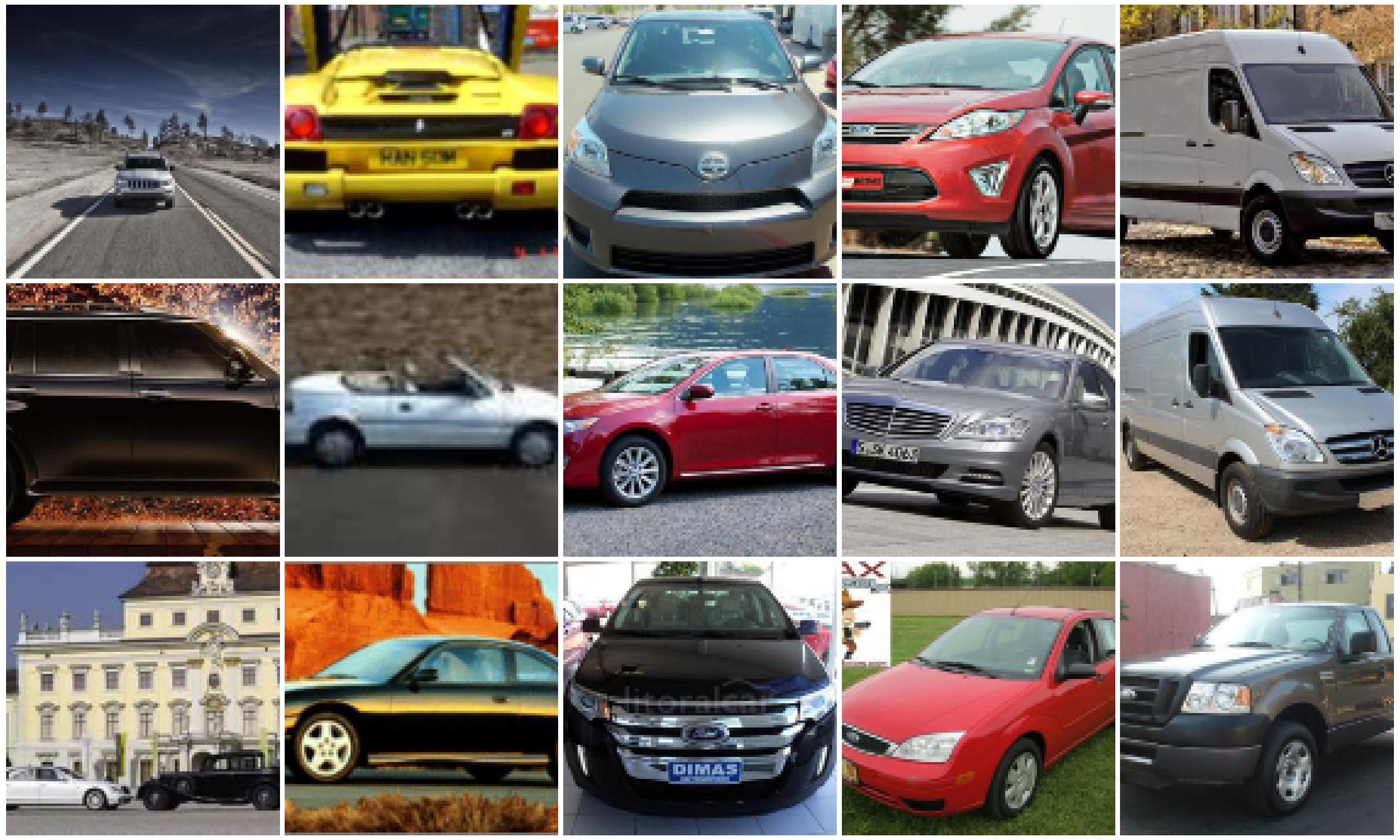}
    \end{minipage}
  \end{subfigure}
  \begin{subfigure}{0.45\linewidth}
    \begin{minipage}{0.05\linewidth}
      \rotatebox{90}{CUB200}
    \end{minipage}
    \begin{minipage}{0.9\linewidth}
        \includegraphics[width=\linewidth]{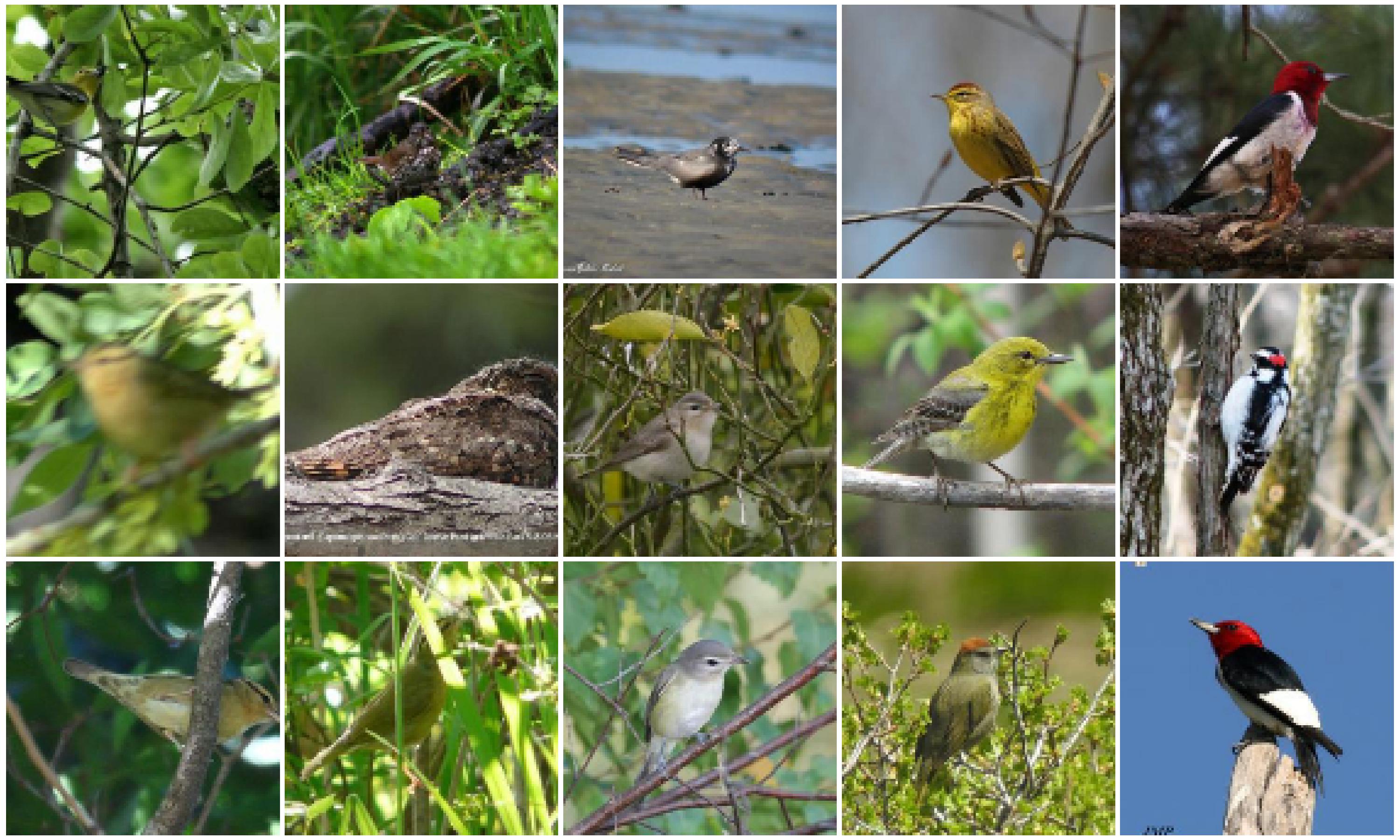}
    \end{minipage}
  \end{subfigure}
  \begin{subfigure}{\linewidth}
  ~\vspace{.1in}
  \end{subfigure}
  \begin{subfigure}{0.45\linewidth}
    \begin{minipage}{0.05\linewidth}
      \rotatebox{90}{In-shop}
    \end{minipage}
    \begin{minipage}{0.9\linewidth}
        \includegraphics[width=\linewidth]{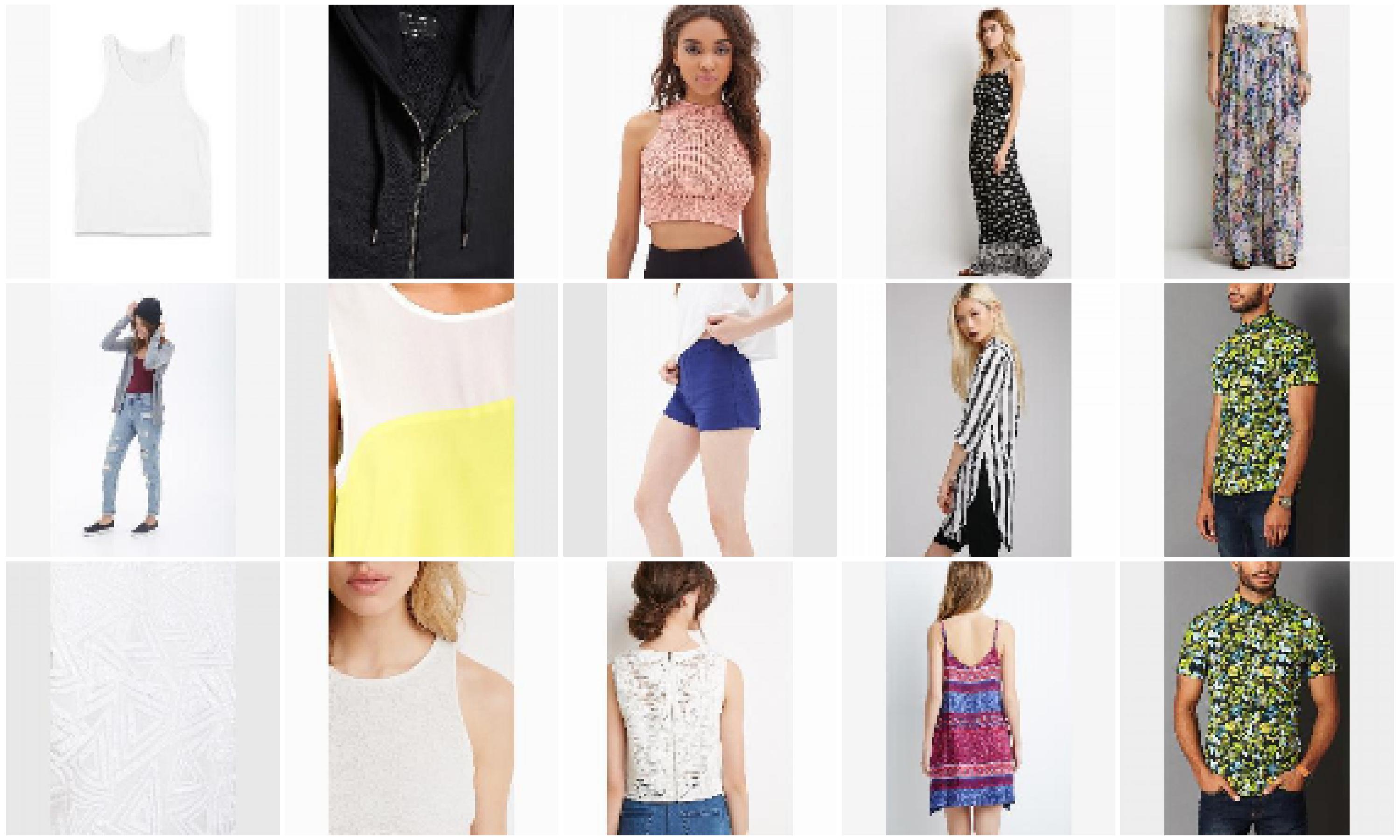}
    \end{minipage}
  \end{subfigure}
  \begin{subfigure}{0.45\linewidth}
    \begin{minipage}{0.05\linewidth}
      \rotatebox{90}{SOP}
    \end{minipage}
    \begin{minipage}{0.9\linewidth}
        \includegraphics[width=\linewidth]{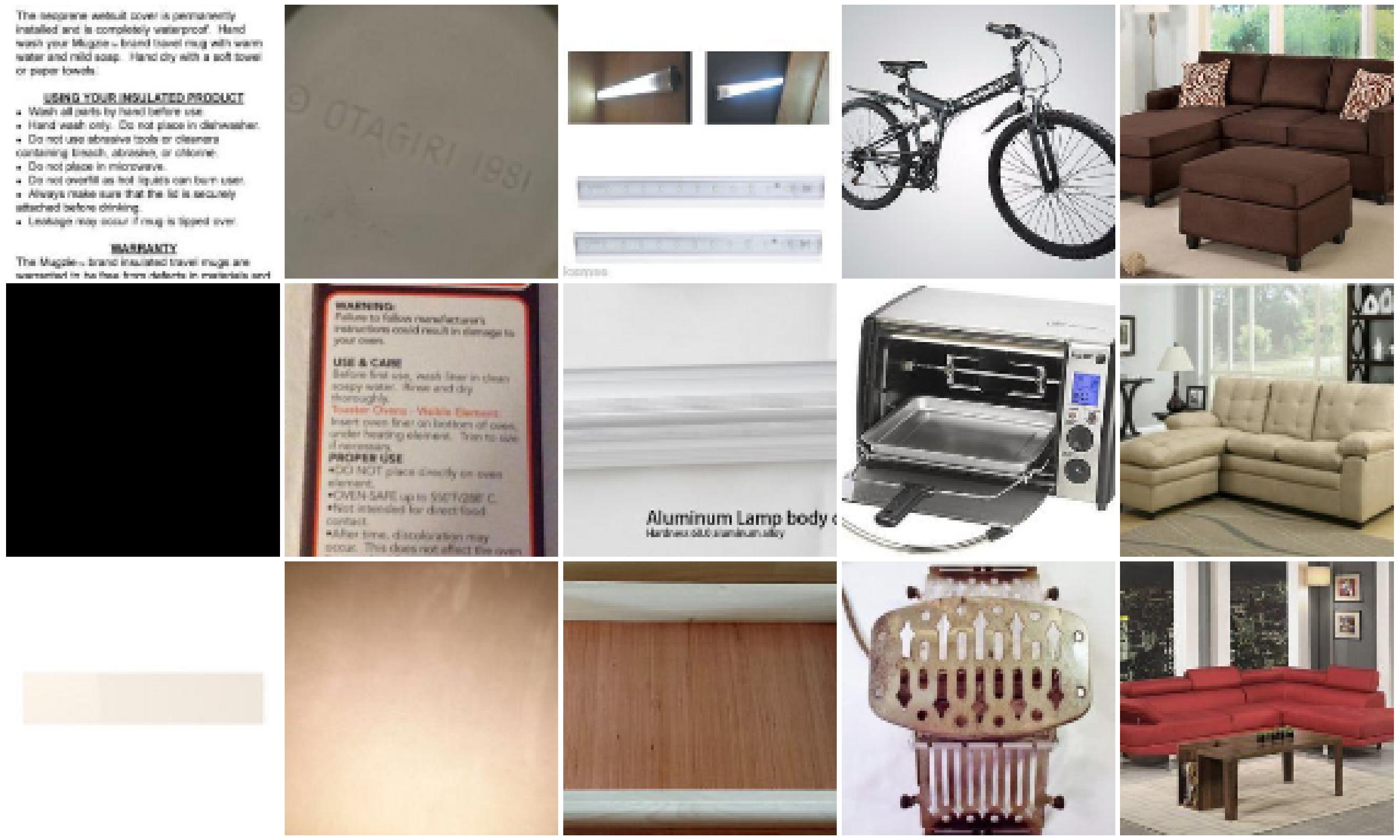}
    \end{minipage}
  \end{subfigure}
  \begin{subfigure}{0.45\linewidth}
    \includegraphics[width=\linewidth]{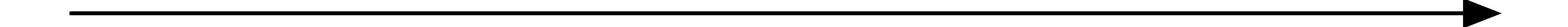}
    \captionsetup{labelformat=empty}
    \caption{Confidence}
  \end{subfigure}
  \begin{subfigure}{0.45\linewidth}
    \includegraphics[width=\linewidth]{tile-arrow.jpg}
    \captionsetup{labelformat=empty}
    \caption{Confidence}
  \end{subfigure}
  \caption{Examples of images with different predicted confidence for DUL-reg-cls method. Confidence increases from left to right.}
  \label{fig:uncertainty}
\end{figure*}

Distribution entropy, predicted by probabilistic methods for each input image, can be used as a measure of model uncertainty \cite{shi2019probabilistic}. The inverse of entropy is sometimes called confidence, and can potentially be used for outlier and adversarial sample detection, dataset cleaning, classification with rejection, active learning and best shot selection \cite{scott2019spe}.
Because of this, we raise the following questions: (1) does confidence score correlate with prediction quality, and (2) can confidence be used for data quality assessment? To answer these questions, we use corrupted variants of the datasets from our evaluation protocol.

As a baseline, we implement two popular approaches to confidence estimation in deterministic models. Classification-based ArcFace and CosFace methods can estimate confidence as a maximum posterior class probability \cite{jung2021standardized}. Some metric learning models implicitly encode confidence to the embeddings' magnitude before the L2-normalization layer \cite{meng2021magface}. The magnitude estimation outperformed maximum posterior in all experiments, presented below.

In our first experiment, we evaluate methods quality for different filter-out rates using the methodologies from previous works on data quality assessment \cite{el2010riskcoverage, li2021spherical}. For a given filter-out rate $\alpha$ and \mbox{dataset size N}, we exclude approximately $N\alpha$ samples with the lowest confidences. We then evaluate the MAP@R metric for the remaining part of the dataset. Corresponding curves for different methods are shown in Figure \ref{filter-out}. It can be seen that curve position largely depends on initial MAP@R value. At the same time, most probabilistic methods outperform deterministic baselines for large filter-out rates even if initial quality had minor or no improvement. In general, PFE and DUL-reg-cls demonstrate the highest quality for most filter-out rates.

In our second experiment, we measure the model's ability to predict retrieval error using confidence by trying to detect cases where the embedding nearest to the query has a different label. For this purpose, we introduce Confidence-based Error Detection Accuracy (CEDA), which is evaluated for the threshold with maximum accuracy. We compute CEDA for training and testing parts of the datasets. As retrieval quality largely depends on the number of classes, we reduce the train set during evaluation to match the size of the test set. Retrieval quality and CEDA are reported in Table \ref{tab:confidence-generalization-train} and Table \ref{tab:confidence-generalization-test}. As a baseline, we use ArcFace for Cars196 and CUB200 datasets and CosFace for In-shop and SOP datasets with confidences evaluated in the same way as in the previous experiment. It can be seen that CEDA is usually close to Recall@1 for all methods. Note that the naive classifier, which always predicts ``no error'', has CEDA equal to Recall@1. We can therefore conclude that current probabilistic methods can safely discard only a small number of errors in data and are incapable of accurate error prediction and classification with rejection. However, we are unable to evaluate the generalization ability of the predicted confidence, as CEDA is close to Recall@1 for both training and testing parts of the dataset.

Finally, we study the ability of probabilistic methods to evaluate data quality. We perform this by computing Spearman's rank correlation coefficient \cite{wu2018iqametrics} between predicted confidence and image crop size on corrupted datasets. Results are presented in Table \ref{tab:lossy-scc}. Confidences from DUL and vMF-FL have the highest correlation among all methods including deterministic baselines. Examples of images with different confidences for DUL-reg-cls are presented in Figure \ref{fig:uncertainty}. It can be seen that images with the lowest confidences are usually blurred, cropped or contain out-of-domain data. Therefore, we suggest using probabilistic methods for data quality assessment in future works.

\section{Future Work}
Top-performing probabilistic methods, including DUL-reg, PFE and SCF, use deterministic pretraining. These methods are usually based on ArcFace and CosFace. On the other hand, Multi-similarity outperforms both methods in terms of verification accuracy. Future work can consider probabilistic extensions of Multi-similarity to further improve verification performance.

In our experiments, we have shown that confidence, predicted by known methods, cannot be directly used for classification with rejection due to low error prediction quality. We thus suggest two directions for future work. First, attention must be paid to confidence's ability to predict verification and retrieval errors, something which was not studied in previous PE works. Second, confidence must generalize well to unseen data. Generalization is hard to estimate for current methods due to low error prediction quality on the training set.

\section{Conclusion}
While many probabilistic methods were proposed for the face verification domain, it was unclear how well they would perform on other tasks. In addition, some essential ablation studies and comparisons between different branches of research were not provided in previous studies. In this work, we compared probabilistic methods in multiple image retrieval and verification tasks. Our experiments show that results depend on the dataset's size. Probabilistic embeddings demonstrate on-par performance with deterministic ones in problems with hundreds of classes. On datasets with thousands of classes, probabilistic embeddings are superior to deterministic baselines in terms of retrieval quality. We also made ablation studies and showed that multivariate normal distribution with distance scoring achieves the highest quality in most cases while having the lowest computational complexity.
Furthermore, probabilistic methods provide out-of-the-box confidence estimation. The predicted confidence correlates with image quality and can be used for data quality assessment. Future research of probabilistic methods can be focused on accurate confidence prediction and confidence generalization for the retrieval with rejection task.

\bibliography{pml}
\bibliographystyle{icml2021}

\end{document}